# UNSUPERVISED SEGMENTATION OF MULTISPECTRAL IMAGES WITH CELLULAR AUTOMATA

**Wuilian Torres**
*wtorres@fii.gob.ve*
Centro de Procesamiento Digital de Imágenes, Instituto de Ingeniería
Urb. Monte Elena II, Sartenejas, Baruta, Caracas-Venezuela.
Laboratorio de Computación Gráfica y Geometría Aplicada, Escuela de Matemática, Facultad de Ciencias, Universidad Central de Venezuela, Los Chaguaramos, Caracas. Venezuela.
**Antonio Rueda Toicen**
*antonio.rueda.toicen@gmail.com*
Centro de Computación Gráfica, Escuela de Computación, Universidad Central de Venezuela, Los Chaguaramos, Caracas. Venezuela

**Abstract**. *Multispectral images acquired by satellites are used to study phenomena on the Earth's surface. Unsupervised classification techniques analyze multispectral image content without considering prior knowledge of the observed terrain; this is done using techniques which group pixels that have similar statistics of digital level distribution in the various image channels. In this paper, we propose a methodology for unsupervised classification based on a deterministic cellular automaton. The automaton is initialized in an unsupervised manner by setting seed cells, selected according to two criteria: to be representative of the spatial distribution of the dominant elements in the image, and to take into account the diversity of spectral signatures in the image. The automaton's evolution is based on an attack rule that is applied simultaneously to all its cells. Among the noteworthy advantages of deterministic cellular automata for multispectral processing of satellite imagery is the consideration of topological information in the image via seed positioning, and the ability to modify the scale of the study. The segmentation algorithm was tested on images acquired by the Venezuelan Miranda satellite.*

## 1. INTRODUCTION

Unsupervised classification of multispectral satellite images facilitates the analysis of the Earth's surface without prior knowledge of the observed terrain; this is done using techniques that group pixels with similar digital level distribution among the different channels of the image. For this task, the most commonly used algorithms are Isodata and K-means [1]. These algorithms perform pixel-based processing and their computation time is dependent on the number of pixels considered, which negatively impacts their response time in images with high spatial resolution. In these images, it is commonly preferred to use object-oriented classification techniques. Spatial segmentation is the first step of object oriented classification, this is done grouping pixels with similar spectral signature; the produced segments characterize homogenous elements in the image, which are then manually pieced together to form an object that's relevant to a particular geographical study. Recent versions of the ENVI image processing software [2] have functionality which allows an interactive performance of this procedure through the manual selection of spectral thresholds. This manual process is time-consuming and prone to operator errors, producing oversegmentation of the multispectral image, or the undesired elimination of segments whose scale is relevant to the scale of the study. We present a methodology of unsupervised segmentation, based in deterministic cellular automata that includes the automatic configuring of its initial state and the elimination of segments with an area inferior to the one that is relevant to the study's scale. We applied the proposed algorithm to the image shown in Figure 1, which depicts fields in Turén, Portuguesa state, Venezuela; the image was acquired by the Miranda satellite, by the MSS2 sensor, on 04/03/2013.

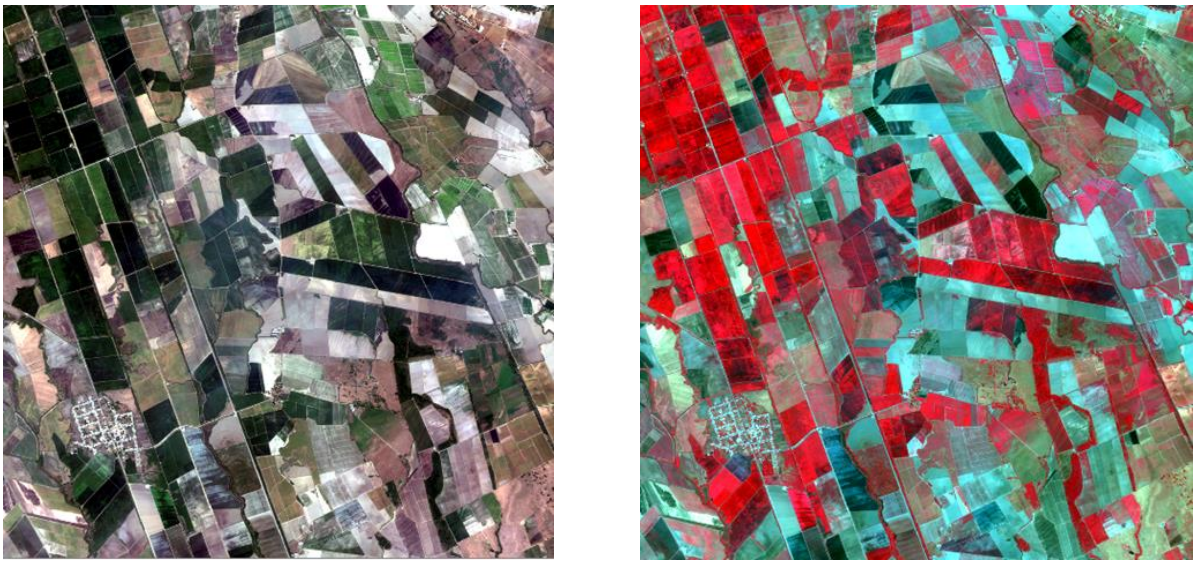

Figure 1 – Multispectral image of Turén, Portuguesa state, Venezuela, acquired by the Miranda satellite. Shown in true color at left and near-infrared color at right.

## 2. DETERMINISTIC CELLULAR AUTOMATA FOR IMAGE SEGMENTATION

A deterministic cellular automaton is a dynamic model composed by an array of cells that evolves through a succession of states *t*. In the automata considered, each one of these cells is in a particular state characterized by a strength level $\theta$ and a label $L$ [3]. At each evolution step, a function is applied simultaneously in all of the automaton's cells. This function evaluates each cell to determine its next state, taking into consideration the current states of its neighboring cells. The state $S_p$, of the currently evaluated cell $p$, evolves under the influence of each of its neighbors $q$, each one characterized by a state $S_q$, as shown in the Algorithm 1. The automaton's evolution can be described as the competitive colonization of the cell space in the image, where each cell is attacked by its neighbors with a strength that is the product of the neighbor's current strength and a term that's inversely proportional to the distance in digital levels between the current cell and its attacking neighbor. The way that attack strength is calculated, makes regions of high homogeneity in digital level likely to be colonized by the same label. We use a modification of the GrowCut evolution rule; it has an attack strength evaluation difference with respect to the original rule that brings a significant reduction in computation time, and a qualitative effect on the obtained segmentation. These differences, and other applicable automata evolution rules, are mentioned in detail in other works by the authors [4], [5].

```
E = true
while (E)
        E = false
        // for each cell in the automaton
        for ∀p ∈ P
                // copy previous state
                L_p^{t+1} = L_p^t
                θ_p^{t+1} = θ_p^t
                // neighbors attack currently evaluated cell
                for ∀q ∈ N(p)
                        if g(‖I_p − I_q‖_2) · θ_q^t > θ_p^{t+1}
                                L_p^{t+1} = L_q^t
                                θ_p^{t+1} = g(‖I_p − I_q‖_2) · θ_q^t
                                E = true
                        end if
                end for
        end for
end while
```

Algorithm 1 – Automaton's Evolution Rule Used in Example

## 3. UNSUPERVISED DEFINITION OF THE AUTOMATON'S INITIAL STATE.

The automaton is initialized with seed cells, which have a label $L$ different from *null* and strength $\vartheta > 0$. Apart from the seed cells, the rest of the automaton's cells have label $L$ equal to *null* and strength $\vartheta = 0$. To achieve a high segmentation quality, seed cells must be selected following 2 criteria:
1) Seed cells must be representative of the spatial distribution of salient elements in the image.
2) Seed cells must be representative of the diversity of spectral signatures present in the image.
    The proposed method for the selection of seed cells does a sampling in the domain of $N$

channels of multispectral images. The N-dimensional space was subdivided into regions according to the distribution of the predominant digital levels in the image. The selection procedure is the following:

1) Pixels are categorized based in the histogram of the multichannel sum of their digital levels, as shown in Figure 2a. This discriminates image regions according to their brightness. The cells with less bright regions, like water bodies, have a relatively low sum, while the brightest, like bare soil, have a relatively high sum. Figure 2b illustrates this discrimination in a bi-dimensional (2 band) case, the blue and red lines represent the spectral levels where the bands $B_1$ and $B_2$ sum the values $P_1$ and $P_2$, respectively.

2) The pixels that share a sum value are now discriminated according to the digital level that dominates each band of the image. In the bi-dimensional case shown in Figure 2b, the region R2 corresponds to pixels with similar digital levels in the 2 bands considered. The region R1 groups pixels where the band 1 has the higher digital level, the region R3 groups pixels where the band 2 has the higher digital level.

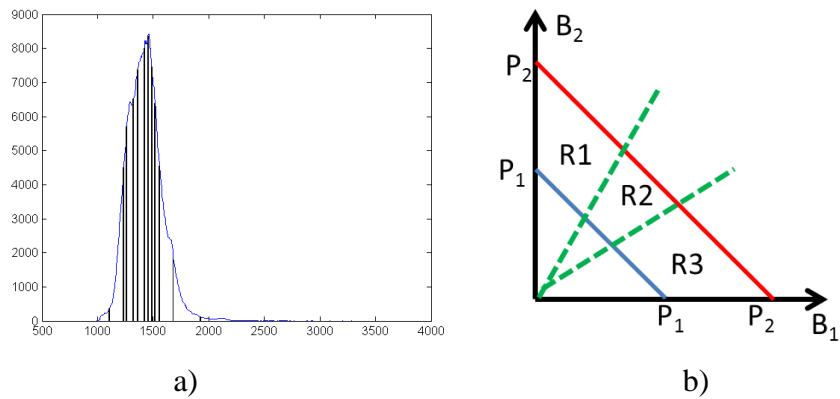

a)  b)

Figure 2 – Seed selection a) Histogram of the sum of bands. The horizontal axis shows the value of the sum of the digital levels in the bands. The vertical axis shows the amount of cells that share the corresponding sum. Black lines indicate the ranges of sum values selected as representative of the digital level distribution. b) Regions of seeds ($R_1$, $R_2$, $R_3$), obtained when considering two bands ($B_1$, $B_2$).

To illustrate the seed selection process in multispectral images, three of the bands of the image presented in Figure 1 were used. As shown in Figure 2a, multi-band digital levels are added and ranges are chosen in a way that is representative of the diverse digital level distribution of the predominant elements in the image, in this example, the ranges correspond to local maxima in the histogram.

At the second phase of the seed selection procedure, we consider the pixels whose sums of bands correspond to the selected ranges. Considering digital levels in three bands, four regions can be identified: the region where the three bands have similar digital levels and another three where one of the bands has a predominant digital level. Figure 4a shows the distribution of seeds in the spectral domain; parallel planes correspond to the sums of components, the pixels on each of these planes are grouped with respect to the predominant band. The localization of seeds in the spatial domain is shown in Figure 4b as dots. Each of these seeds receives a label according to the region where its spectral signature is located; in the example in the Figure 4b, the seeds are distributed in 18 spectral regions, corresponding to 18 labels.

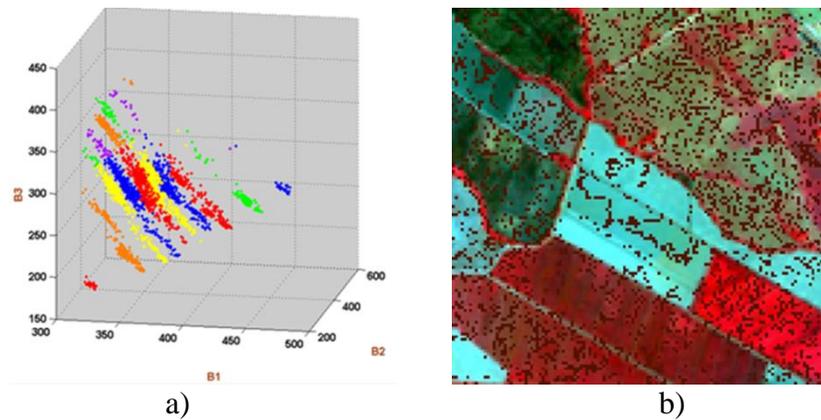

a) b)

Figure 4- a) Distribution of the seeds in the spectral domain of the image. b) Localization of the seeds in the spatial domain of the image.

Figure 5a shows the seeds obtained from the processing of the multispectral image of Turén shown in Figure 1. In this example, 101175 seeds, corresponding to 10.1% of the pixels in the image, where assigned 46 distinct labels. The image is segmented when the automaton converges after 91 evolution steps. Figure 5b shows the final label distribution; the segmented image has 27171 segments, a significant amount of these segments has individual area that's below a desired scale of study of 150 pixels. To eliminate over-segmentation, the automaton's evolution is reinitiated using the following procedure:

1) The segments with area inferior to the desired scale of the study are identified; in this example we select those that have less than 150 pixels.
2) The state of the cells in these segments is modified reassigning the cell's existing labels to *null* and their strengths to 0.
3) The automaton's evolution is reinitiated from this state.

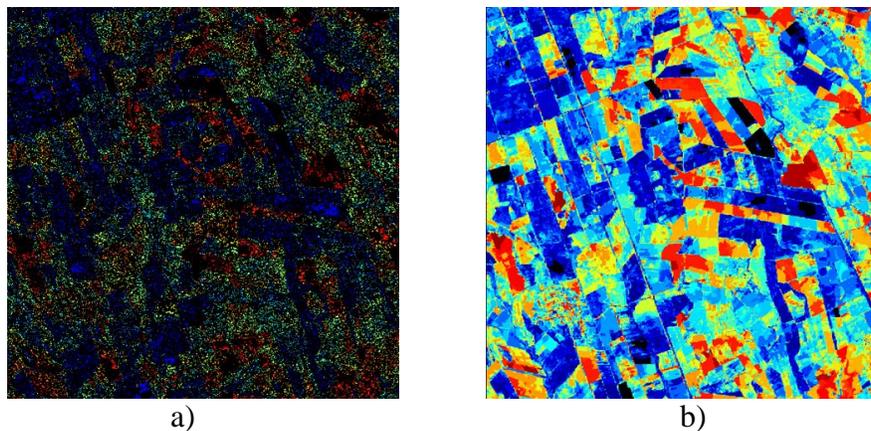

a) b)

Figure 5- a) Initial automaton's configuration ($t = 0$) showing seed positioning b) Segmentation at $t = 91$.

## 4. ASSIGNMENT OF SPECTRAL SIGNATURE

With the purpose of using the segmentation result obtained as input for an object-oriented classification process, each segment receives a single spectral signature to its contained cells. Each cell has an N-dimensional vector that is composed by the digital levels of every band considered. The spectral signature of a segment is obtained by finding the centroid of the N-dimensional vectors formed by the digital levels in the group of pixels in the segment. The centroid is the vector positioned at the least distance to all the other vectors in the segment. Figure 6a shows the vector *a* as the representative centroid of the pixels *a*, *b*, *c*, *d*, *e*.

Figure 6b shows the segmented image obtained, this time with 1593 segments. To ease its interpretation, a color palette was used where each segment contains the average color of the false color in near infrared shown in the Figure 1.

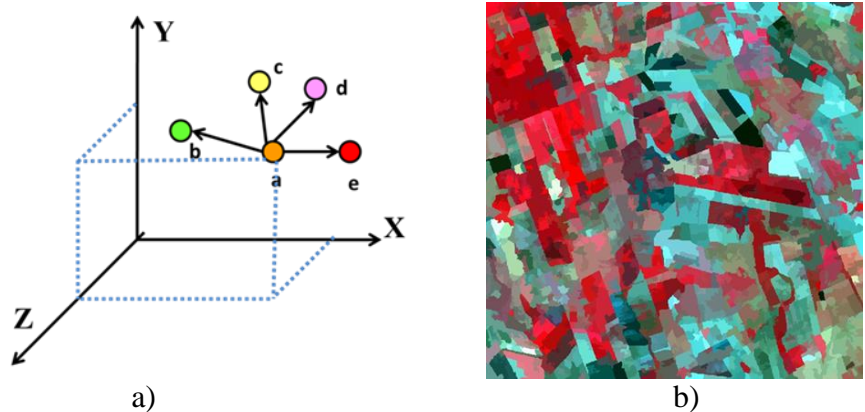

a)          b)

Figure 6- a) Centroid, the spectral signature *a* becomes the representative signature for its segment. b) Segmented image using centroids as colors, the segments with area inferior to 150 cells have been eliminated through competitive evolution.

## 5. CONCLUSIONS

Unsupervised segmentation through deterministic cellular automata, using seeds determined via an analysis of multispectral information, is presented as a powerful tool to segment images of high spatial resolution. Commercially available systems that provide object-oriented classification functionality require manual participation of the user to determine thresholds that reduce the over-segmentation of the image and reduce the loss of segments that are important for the particular study. Our proposed methodology allows the automatic creation of segments that are appropriate for the scale of the study, reducing necessary user interaction